\documentclass[11pt]{article}

\usepackage[preprint]{acl}

\usepackage{times}
\usepackage{latexsym}

\usepackage[T1]{fontenc}

\usepackage[utf8]{inputenc}

\usepackage{microtype}

\usepackage{inconsolata}

\usepackage{graphicx}
\usepackage{subcaption}
\usepackage{colortbl}
\usepackage{hyperref}
\usepackage{url}
\usepackage{booktabs}
\usepackage{tabularx}
\usepackage{amssymb}
\usepackage{arydshln}
\usepackage{multirow}
\usepackage{graphicx}
\usepackage[svgnames]{xcolor}
\usepackage[misc]{ifsym}
\usepackage{wrapfig}
\usepackage{subcaption}
\usepackage{enumitem}
\usepackage{algorithm}
\usepackage{algpseudocode}
\usepackage{fontawesome}
\usepackage{hyperref}
\usepackage{amsmath}

\title{Seer Self-Consistency:
Advance Budget Estimation for \\ Adaptive Test-Time Scaling}

\author{%
  Shiyu Ji\footnotemark[2]\quad{Yixuan Wang}\footnotemark[2]\quad{Yijun Liu}\quad\textbf{Qingfu Zhu}\quad\textbf{Wanxiang Che}\footnotemark[1] \\
  Research Center for Social Computing and Interactive Robotics, \\
  Harbin Institute of Technology, China \\
  \texttt{\{syji,car\}@ir.hit.edu.cn} \\
  \faGithubAlt~\url{https://github.com/noforit/SeerSC}
}

\begin{document}
\maketitle

\renewcommand{\thefootnote}{\fnsymbol{footnote}}
\footnotetext[2]{Equal contribution.}
\footnotetext[1]{Corresponding author.}
\renewcommand{\thefootnote}{\arabic{footnote}}

\begin{abstract}
Test-time scaling improves the inference performance of Large Language Models (LLMs)
but also incurs substantial computational costs.
Although recent studies have reduced token consumption through dynamic self-consistency, they remain constrained by the high latency of sequential requests.
In this paper, we propose SeerSC,
a dynamic self-consistency framework that simultaneously improves token efficiency and latency
by integrating System 1 and System 2 reasoning.
Specifically, we utilize the rapid System 1 to compute the answer entropy for given queries.
This score is then used to evaluate the potential of samples for scaling,
enabling dynamic self-consistency under System 2.
Benefiting from the advance and accurate estimation provided by System 1,
the proposed method can reduce token usage while simultaneously achieving a significant decrease in latency through parallel generation.
It outperforms existing methods, achieving up to a 47\% reduction in token consumption and a 43\% reduction in inference latency without significant performance loss.

\end{abstract}

\section{Introduction}
Large Language Models (LLMs) \citep{hurst2024gpt,yang2024qwen2,guo2025deepseek}
have demonstrated exceptional performance in reasoning tasks,
particularly in mathematics.
By leveraging methods such as self-consistency \citep{wang2022self},
LLMs can substantially improve their performance during inference
through increased computation \citep{chen2025towards,zhang2025survey}
without additional training.
While improving performance,
Test-Time Scaling (TTS) methods also significantly
increase the computational cost \citep{brown2024large,chen2024more}.
Many efforts \citep{feng2025efficient,liu2025efficient} focus on
optimizing the trade-off between performance and efficiency.

\begin{figure}[t]
    \centering
    \includegraphics[width=0.9\linewidth]{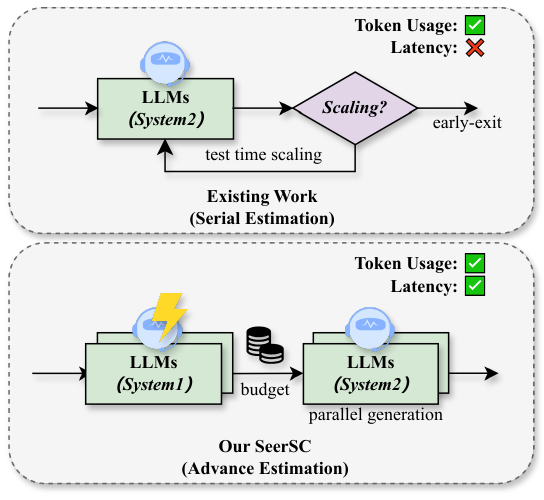}
    \caption{Comparison of the Proposed method with Existing Methods.
Unlike sequential estimation methods,
SeerSC leverages fast System 1 to guide the parallel generation of System 2.}
    \label{fig:intro}
\end{figure}

\begin{figure*}[t]
    \centering
    \subcaptionbox{Test-Time Scaling is not beneficial for every sample.
    Employing a dynamic budget allocation is necessary.
    \label{fig:obs1}}
   {\includegraphics[width=0.32\textwidth]{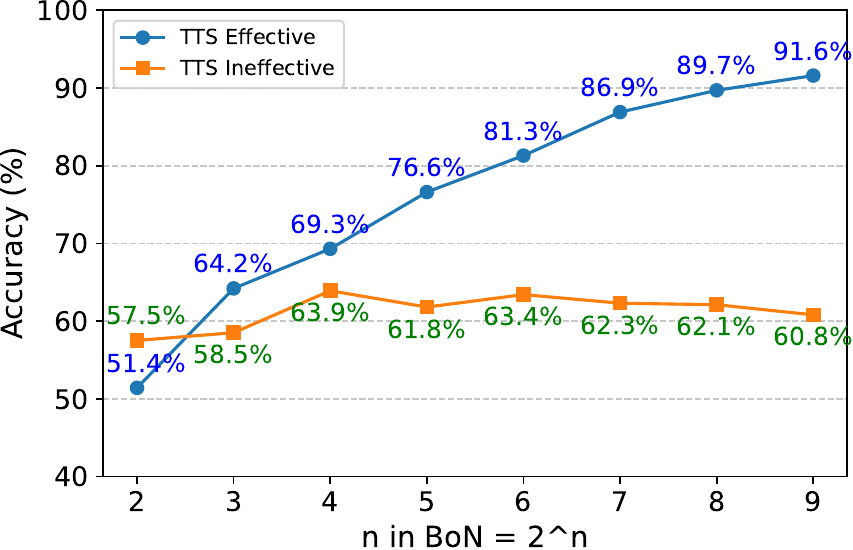}}
    \hfill
   \subcaptionbox{The diversity of generated answers can reflect the potential for scaling.\label{fig:obs2}}
   {\includegraphics[width=0.32\textwidth]{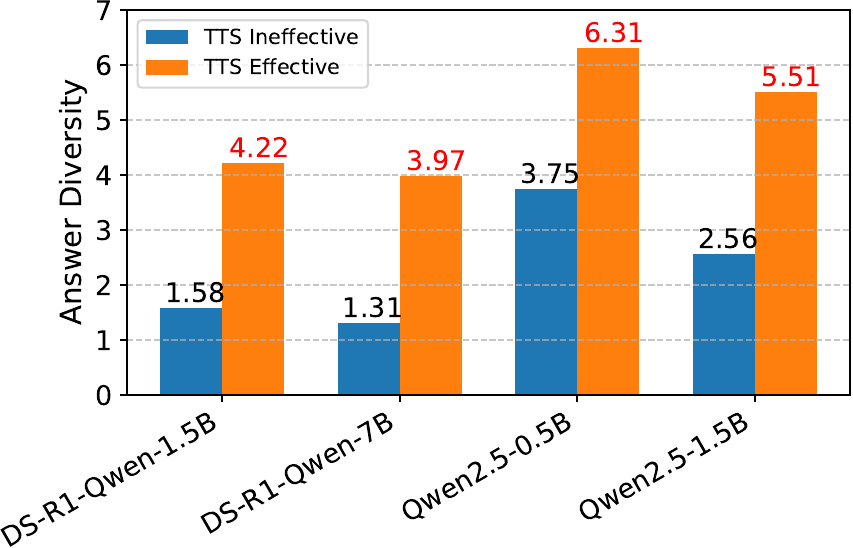}}
    \hfill
   \subcaptionbox{The answer entropy of early-stopped generation exhibits consistency across varying reasoning steps.\label{fig:obs3}}
   {\includegraphics[width=0.32\textwidth]{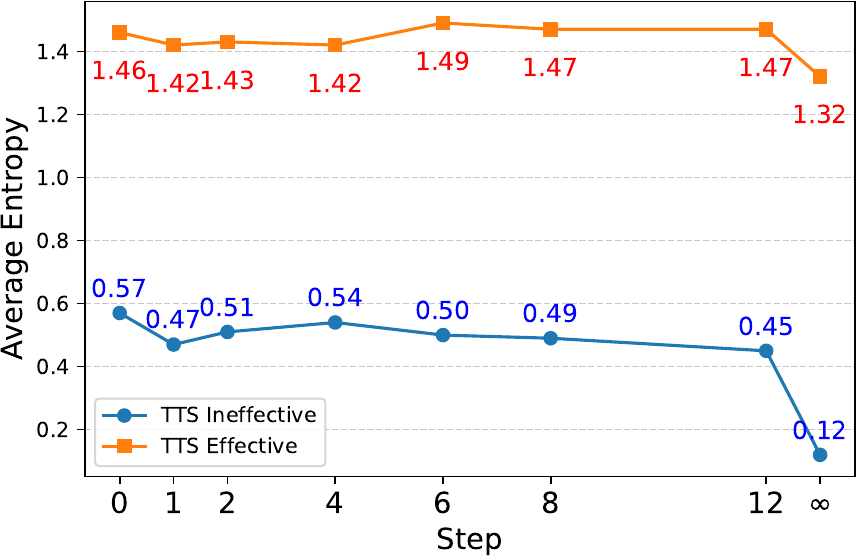}}
    \caption{Observational Experiments on the Math Dataset.}
    \label{fig:placeholder}
\end{figure*}

For ensemble methods like Self-Consistency (SC),
a critical challenge is dynamically determining the optimal number of ensemble samples based on the complexity of the input instance \citep{wang2024make}.
Adaptive Consistency (AC) \citep{aggarwal2023let} generates responses sequentially and decides whether to terminate scaling based on the diversity of the current answers and a specified threshold.
Early-Stopping Self-Consistency (ESC) \citep{liescape} employs a sliding window to track the diversity of the current responses, enabling dynamic early stopping for self-consistency.
By dynamically adjusting the sampling budget,
these methods significantly reduce token consumption during the TTS phase.

However, a reduction in token usage does not always translate into a corresponding decrease in latency \citep{wang2025faster,huang2025latency}.
Existing dynamic SC methods typically estimate the required computation for each query through \textbf{sequential requests},
as shown in Figure \ref{fig:intro},
which \textbf{conflicts with the efficient parallel sampling techniques} used by modern inference engines \citep{kwon2023efficient,zheng2024sglang}.
This issue is further exacerbated for recent reasoning models \citep{guo2025deepseek} as the output length increases.
If a computational budget can be allocated to different samples prior to their execution,
we can leverage parallel sampling to concurrently optimize both the token consumption and the inference latency, leading to more efficient TTS.

In this paper, we propose \textit{SeerSC}, a dynamic self-consistency framework that integrates intuition (System 1) and reasoning (System 2) \citep{li2025system} to enable the pre-allocation of computational resources.
Observing the consistency in answer diversity between the rapid System 1 and deliberative System 2 generation forms,
we propose using the entropy of the fast System 1 answers to guide the computational budget allocation for System 2.
Specifically, we first use the System 1 approach to generate multiple responses for the sample in parallel, and subsequently calculate the entropy of the resulting answers.
Based on the answer entropy score, we can achieve fine-grained budget allocation, directing limited computational resources to
samples with greater scaling potential to maximize performance improvement.

We conduct experiments across multiple model configurations, including reasoning and instruct settings, and utilize mainstream mathematical reasoning benchmarks.
By leveraging the advance budget estimation from the rapid System 1,
SeerSC is able to decrease the token usage during the TTS stage
while exploiting parallel generation to mitigate overall latency.
Our method demonstrably surpasses current techniques, delivering up
to 47\% lower token usage and 43\% reduced inference latency
without substantial performance degradation.
Furthermore, compared to sequential methods,
our approach can be more effectively integrated with path-pruning techniques
\citep{fu2025deep},
achieving orthogonal performance gains and expanding its range of application.

Our main contributions can be summarized as
follows:
\begin{itemize}
    \item We propose Seer Self-Consistency (SeerSC),
    which utilizes the answer entropy of System 1 to guide the dynamic budget allocation of System 2 during TTS stage.
    \item We propose a fine-grained budget allocation method that ensures computational resources are effectively directed to samples exhibiting a high potential for scaling.
    \item Experiments conducted across multiple settings demonstrate that the proposed method outperforms existing dynamic SC approaches in both token consumption and latency. 
\end{itemize}

\begin{figure*}[t]
    \centering
    \includegraphics[width=0.9\textwidth]{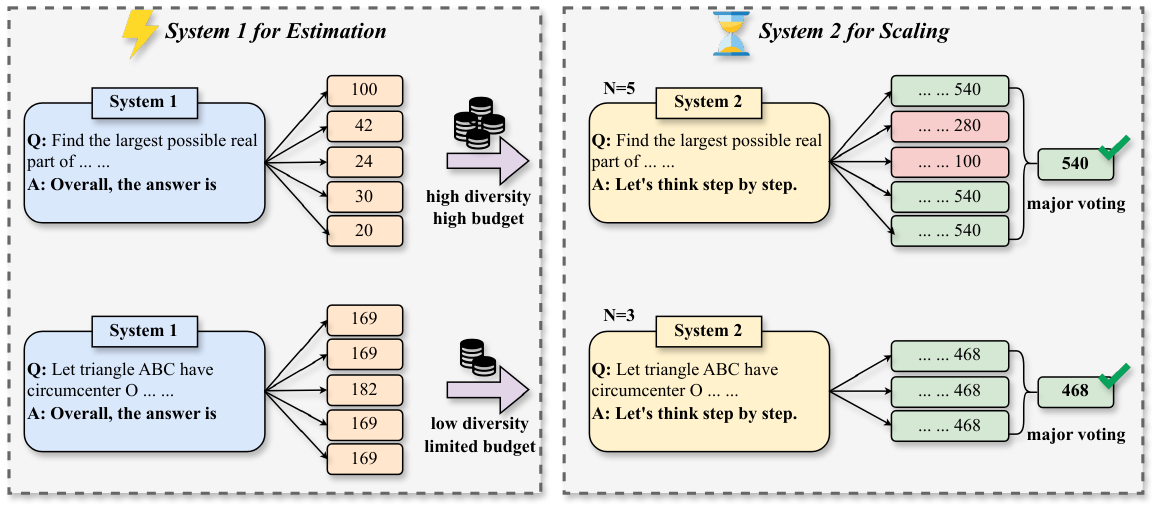}
    \caption{Overall workflow of the proposed SeerSC framework.
We first utilize System 1 to rapidly estimate the answer diversity for each sample,
and subsequently allocate varying computational resources for the System 2 phase based on the answer entropy.}
    \label{fig:main}
\end{figure*}

\section{Observation}
\label{sec:obs}
As discussed, achieving parallel inference with a dynamic budget requires determining the budget for each sample prior to execution.
In search of an effective metric for budget allocation,
we initiate an analytical investigation into the gains provided by self-consistency. Our findings resulted in the following two primary observations:

\paragraph{Answer diversity can reflect the scaling potential.}
As shown in Figure \ref{fig:obs1}, 
not all samples benefit from Test-Time Scaling (TTS).
Thus, an effective metric is required to guide budget allocation.
We further perform a statistical analysis of the number of unique answers resulting from Self-Consistency.
As shown in Figure \ref{fig:obs2},
samples with negligible TTS gain typically yielded a lower number of answers, indicating a higher degree of certainty in the responses.
This insight inspires us to use the entropy of the generated answers as a metric to determine the TTS budget for each sample.

\paragraph{The answer entropy of System 1 and System 2 exhibits consistency.}
Nevertheless, acquiring the answer entropy before response generation is unfeasible. While current approaches \citep{aggarwal2023let,liescape} compute the entropy via serial execution, they still suffer from significant latency challenges.
In order to compute the certainty of various samples in advance,
we investigate the use of low-cost techniques for approximating the answer entropy.
As shown in Figure \ref{fig:obs3}, we illustrate the answer entropy of samples generated via early stopping after a varying number of steps.
The leftmost point represents System 1, which generates the answer directly, while the rightmost point represents System 2, which involves full reasoning.
As evident from the figure, the answer entropy for the two classes of samples (those that benefit from TTS and those that do not) shows a distinct difference in both System 1 and System 2.
This confirms the potential for utilizing the rapid System 1 process to estimate the answer entropy obtained from System 2.

\section{Methodology}
\subsection{Overall}
Based on these observations,
we propose Seer Self-Consistency (SeerSC), a dynamic budget Test-Time Scaling framework that integrates the intuition of System 1 and the reasoning of System 2.
The overall workflow of the framework is illustrated in Figure \ref{fig:main}.
We first perform System 1-level inference to obtain the answer entropy (\S \ref{sec:sys1})
for each sample.
Subsequently, we rank the samples based on their answer entropy to
allocate differential amounts of computation (\S \ref{sec:sys2}) according to the pre-defined budget.
This allows us to efficiently execute System 2-level TTS within a given budget.

\subsection{System 1 for Estimation}
\label{sec:sys1}



As illustrated in Figure~\ref{fig:main}, given a problem $X$, 
System~1 first performs direct answer generation (see Appendix~\ref{sec:system1_implementation} for implementation details on reasoning models) with a sampling size of $M$, 
producing an answer set $\mathrm{Ans} = \{A_1, A_2, \dots, A_M\}$.

For each answer $A \in \mathrm{Ans}$, we define a confidence score $S(x)$ based on its token-level probabilities.  
Let $A = [a_1, a_2, \dots, a_k]$ where $k = |A|$ and $a_i$ denotes the $i$-th token in the answer.  
The confidence is defined as:
\begin{equation}
    S(A) = \exp\!\left(\frac{1}{k}\sum_{i=1}^k \log p(a_i)\right)
    \label{eq:confidence}
\end{equation}
where $p(a_i)$ denotes the probability of token $a_i$. 
This formulation corresponds to the exponentiated average log-likelihood of the tokens in $A$.

We then compute the entropy $E(X)$ over \emph{answer categories} to guide the load allocation in System~2.  
Unlike vanilla Shannon entropy that depends on unweighted category frequencies, we integrate category confidence weights into the entropy calculation.

Let $\mathcal{C}=\{c_1,\dots,c_m\}$ be the set of distinct answer categories and 
$\mathrm{Ans}=\{A_1,\dots,A_M\}$ the sampled answers. 
The (unnormalized) weight of category $c_j$ is the sum of member-answer confidences:
\begin{equation}
    W(c_j) \;=\; \sum_{A \in \mathrm{Ans},\, A \in c_j} S(A)
    \label{eq:category_weight}
\end{equation}
We normalize to obtain a probability distribution over categories:
\begin{equation}
    \tilde{W}(c_j) \;=\; \frac{W(c_j)}{\sum_{c \in \mathcal{C}} W(c)}
    \label{eq:normalized_weight}
\end{equation}
The confidence-weighted entropy of problem $X$ is then
\begin{equation}
    E(X) \;=\; -\sum_{c \in \mathcal{C}} \tilde{W}(c)\,\log \tilde{W}(c)
    \label{eq:weighted_entropy}
\end{equation}


\subsection{System 2 for Scaling}
\label{sec:sys2}

In System~2, we allocate a computation budget $B(X)$ for each problem $X$ based on its estimated entropy $E(X)$ from System~1. 
As observed in Section~\ref{sec:obs}, problems with lower entropy require fewer sampled reasoning paths, 
whereas problems with higher entropy benefit from a larger number of samples. 
In contrast to Self-Consistency (SC), which assigns a fixed budget of $N$ samples to every problem, 
our adaptive allocation follows the rule:
\begin{equation}
B(X) = 
\left\{
\begin{aligned}
    1,     & \quad E(X) < \tau_1, \\
    N/2,   & \quad \tau_1 \leq E(X) < \tau_2, \\
    N,     & \quad \text{otherwise}.
\end{aligned}
\right.
\label{eq:budget}
\end{equation}
where $\tau_1$ and $\tau_2$ are entropy thresholds.

By definition, the entropy $E(X)$ lies in the range
\begin{equation}
0 \;\leq\; E(X) \;\leq\; \log M
\end{equation}
where $M$ is the number of sampled direct answers in System~1. 
The lower bound is attained when all direct answers fall into a single category, yielding $E(X)=0$. 
The upper bound is attained when every sampled direct answer belongs to a distinct category, i.e., $m = M$, 
where $m$ is the number of unique categories. 
In this case, the category distribution is uniform, and the entropy reaches its maximum value $\log M$.

Based on this range, we empirically set the thresholds as fractions of $\log M$: 
\begin{equation}
    \tau_1 = \tfrac{1}{10}\log M, \quad \tau_2 = \tfrac{1}{3}\log M.
\end{equation}
This empirical setting provides a practical balance between efficiency and accuracy (see Appendix~\ref{sec:sensitivity} for a detailed sensitivity analysis).


\section{Experiment}

\subsection{Experimental Setup}

\paragraph{Evaluation benchmarks and models.}
We conduct evaluations on the following datasets. 
MATH500 \citep{lightman2023let} represents a representative subset of problems from the MATH dataset \citep{hendrycks2021measuring}. 
AIME2024 \citep{aime2024}, AIME2025 \citep{aime2025}, and AMC2023 \citep{amc2023} consist of high-difficulty mathematical competition problems. 
GPQA Diamond \citep{rein2024gpqa} serves as a benchmark for graduate-level STEM reasoning tasks, providing a challenging evaluation setting.
We evaluate on the following models: DeepSeek-R1-Distill-Qwen-7B \citep{guo2025deepseek}, DeepSeek-R1-Distill-Llama-8B \citep{guo2025deepseek}, and Qwen3-4B \citep{yang2025qwen3}. The first two are representative reasoning models, while Qwen3-4B supports both \emph{thinking} and \emph{non-thinking} modes, and we conduct evaluations under both settings.
Additionally, we extend our evaluation to the larger-scale QwQ-32B \citep{qwq32b} model to verify generalizability (see Appendix~\ref{sec:large_models}).

\paragraph{Selected baselines.}
We compare our method against four baseline methods: 
\begin{itemize}
\item \textbf{Chain of Thought (CoT)} \citep{wei2022chain} employs standard CoT prompting, where the model generates a single reasoning trajectory that directly leads to the final answer. 
\item \textbf{Self-Consistency (SC)} \citep{wang2022self} generates $N$ independent reasoning paths and determines the final answer by majority voting on the predicted solutions. 
\item \textbf{Adaptive Consistency (AC)} \citep{aggarwal2023let} sequentially generates reasoning paths in a sequential manner. In each step, it computes the relative frequency of unique answers and terminates sampling once the frequency of a candidate exceeds a predefined threshold or the maximum $N$ is reached. The final answer is again decided by majority voting.
\item \textbf{Early-Stopping Self-Consistency (ESC)} \citep{liescape} improves standard SC by sequentially generating reasoning paths until the upper limit of $N$ is reached or the solutions within a sliding window of size $W$ converge to the same answer. The final prediction is then determined via majority voting. 
\end{itemize}

\begin{table*}[t]
\scriptsize
\centering
\caption{Main experimental results of SeerSC compared with baseline methods (CoT, SC, AC, and ESC) 
across diverse models and benchmarks. 
We report accuracy (\%), average tokens ($\times 10^3$), and average latency (s).}
\label{tab:main_results}
\begin{tabularx}{\textwidth}{c*{15}{>{\centering\arraybackslash}X}}

\toprule
\multirow{2}{*}{\textbf{Methods}}
& \multicolumn{3}{c}{\textbf{MATH500}} 
& \multicolumn{3}{c}{\textbf{AIME2024}} 
& \multicolumn{3}{c}{\textbf{AIME2025}} 
& \multicolumn{3}{c}{\textbf{AMC2023}} 
& \multicolumn{3}{c}{\textbf{GPQA-D}} \\
\cmidrule(lr){2-4} \cmidrule(lr){5-7} \cmidrule(lr){8-10} \cmidrule(lr){11-13} \cmidrule(lr){14-16}
& \textit{Acc.}$\uparrow$ & \textit{Token}$\downarrow$ & \textit{Lat.$\downarrow$}
& \textit{Acc.}$\uparrow$ & \textit{Token}$\downarrow$ & \textit{Lat.$\downarrow$}
& \textit{Acc.}$\uparrow$ & \textit{Token}$\downarrow$ & \textit{Lat.$\downarrow$}
& \textit{Acc.}$\uparrow$ & \textit{Token}$\downarrow$ & \textit{Lat.$\downarrow$}
& \textit{Acc.}$\uparrow$ & \textit{Token}$\downarrow$ & \textit{Lat.$\downarrow$} \\
\midrule

\rowcolor{gray!10}
\multicolumn{16}{c}{\textit{DeepSeek-R1-Distill-Qwen-7B}} \\
\midrule
CoT & 90.0 & -- & -- & 49.7 & -- & -- & 38.8 & -- & -- & 83.9 & -- & -- & 44.7 & -- & -- \\
\hdashline
SC  & \textbf{93.9} & 29.8 & 8.4 & \textbf{74.4} & 344.4 & 158.6 & 58.6 & 368.6 & 174.7 & \textbf{91.6} & 42.5 & 13.2 & 48.1 & 128.5 & 51.8 \\
AC  & 93.8 & 19.4 & 12.4 & 73.7 & \textbf{223.7} & 214.1 & 59.3 & \textbf{286.9} & 233.7 & 91.3 & \textbf{32.8} & 24.8 & 47.7 & 98.3 & 73.1 \\
ESC & 93.6 & 22.2 & 13.9 & 74.0 & 270.4 & 261.2 & 58.3 & 319.5 & 266.8 & 91.1 & 34.9 & 26.7 & 48.0 & 105.9 & 82.0 \\
\rowcolor{blue!10}
SeerSC & 93.0 & \textbf{17.7} & \textbf{5.3}  & 74.1 & 315.6 & \textbf{151.8} & \textbf{60.2} & 356.0 & \textbf{173.8} & 91.5 & 33.5 & \textbf{9.4}  & \textbf{48.6} & \textbf{95.3} & \textbf{37.8} \\
\midrule

\rowcolor{gray!10}
\multicolumn{16}{c}{\textit{DeepSeek-R1-Distill-Llama-8B}} \\
\midrule
CoT & 84.5 & -- & -- & 39.3 & -- & -- & 27.8 & -- & -- & 78.6 & -- & -- & 46.2 & -- & -- \\
\hdashline
SC  & 91.5 & 32.8 & 16.9 & \textbf{68.8} & 379.7 & 376.0 & 44.3 & 387.0 & 395.1 & 88.6 & 45.6 & 28.1 & 50.7 & 74.6 & 62.5 \\
AC  & 91.3 & 24.3 & 21.8 & 67.7 & 290.0 & 361.4 & 44.8 & 312.2 & 392.3 & 88.6 & 34.5 & 39.5 & 49.9 & 62.7 & 71.6 \\
ESC & \textbf{91.7} & 26.7 & 23.3 & 67.9 & 337.3 & 436.6 & \textbf{44.9} & 339.1 & 424.3 & \textbf{88.9} & 37.5 & 40.9 & 50.4 & 66.2 & 75.1 \\
\rowcolor{blue!10}
SeerSC & 90.2 & \textbf{21.2} & \textbf{10.6} & 66.3 & \textbf{288.6} & \textbf{294.0} & 43.8 & \textbf{308.3} & \textbf{300.6} & 87.3 & \textbf{31.5} & \textbf{16.6} & \textbf{50.8} & \textbf{53.0} & \textbf{43.5} \\
\midrule

\rowcolor{gray!10}
\multicolumn{16}{c}{\textit{Qwen3-4B (thinking)}} \\
\midrule
CoT & 88.5 & -- & -- & 60.9 & -- & -- & 51.0 & -- & -- & 78.8 & -- & -- & 59.2 & -- & -- \\
\hdashline
SC  & \textbf{91.7} & 36.5 & 17.7 & \textbf{71.4} & 191.3 & 176.4 & \textbf{65.3} & 209.9 & 205.2 & \textbf{87.0} & 53.3 & 61.7 & 63.0 & 70.8 & 54.8 \\
AC  & 91.6 & 23.8 & 19.0 & 71.2 & \textbf{126.8} & 165.2 & 64.9 & \textbf{151.5} & 197.9 & 86.4 & 39.5 & 54.5 & 62.5 & 53.2 & 56.0 \\
ESC & \textbf{91.7} & 27.0 & 21.5 & 71.3 & 138.4 & 186.6 & 65.2 & 168.5 & 222.6 & 86.9 & 43.1 & 62.5 & \textbf{63.4} & 58.8 & 62.1 \\
\rowcolor{blue!10}
SeerSC & 90.6 & \textbf{19.1} & \textbf{9.5}  & 70.7 & 156.8 & \textbf{142.2} & 64.3 & 183.7 & \textbf{178.2} & 84.8 & \textbf{32.8} & \textbf{35.7} & 62.5 & \textbf{37.6} & \textbf{29.8} \\
\midrule

\rowcolor{gray!10}
\multicolumn{16}{c}{\textit{Qwen3-4B (non-thinking)}} \\
\midrule
CoT & 83.3 & -- & -- & 23.0 & -- & -- & 18.7 & -- & -- & 67.2 & -- & -- & 48.7 & -- & -- \\
\hdashline
SC  & \textbf{89.1} & 7.2  & 1.8 & 39.1 & 98.2  & 36.0 & 22.6 & 49.2  & 17.3 & 76.8 & 13.3 & 4.0 & \textbf{57.6} & 24.4 & 6.5 \\
AC  & 89.0 & 5.5  & 3.1 & 39.2 & \textbf{81.3}  & 59.9 & 22.2 & \textbf{44.5}  & 32.8 & \textbf{76.9} & 11.5 & 7.8 & 57.3 & 16.2 & 11.3 \\
ESC & 89.0 & 6.0  & 3.6 & \textbf{39.8} & 85.6  & 68.4 & 22.4 & 46.7  & 34.7 & 76.6 & 12.0 & 8.0 & \textbf{57.6} & 18.2 & 11.8 \\
\rowcolor{blue!10}
SeerSC & 88.1 & \textbf{4.4}  & \textbf{1.5} & 38.9 & 93.6  & \textbf{33.6} & \textbf{22.9} & 47.8  & \textbf{16.1} & 76.3 & \textbf{9.5}  & \textbf{3.3} & 57.1 & \textbf{13.0} & \textbf{3.7} \\
\bottomrule
\end{tabularx}
\end{table*}

\paragraph{Implementation details.}
To ensure a fair comparison, we search for the optimal sampling number $N$ for the standard SC baseline on each model and dataset, as gains beyond this point are limited due to diminishing returns \citep{chen2024more}. All baselines and our method are evaluated under this $N$ setting. For AC baseline, the stopping threshold is set to $0.95$. For ESC baseline, we follow the original paper and set the window size $W=5$.  For DeepSeek-R1-Distill-Qwen-7B, DeepSeek-R1-Distill-Llama-8B, and Qwen3-4B (thinking mode), we set the maximum output length to $10{,}240$ on Math500 and AMC2023, and to $16{,}384$ on the other datasets. For Qwen3-4B (non-thinking mode), the maximum output length is fixed to $8{,}192$ across all datasets. All experiments are conducted using \texttt{vLLM} on NVIDIA A100-80GB GPUs with CUDA~12.4.
The setting of $N$ and additional implementation details can be found in Appendix~\ref{sec:details}.

\subsection{Main Results}
As shown in Table \ref{tab:main_results}, we report the results of SeerSC and other baseline methods across all evaluated models and datasets. The reported metrics include accuracy, the average number of output tokens and latency per problem, with all experiments repeated multiple times to ensure stability.

\paragraph{SeerSC matches baselines in accuracy.}
Across all models and datasets, SeerSC consistently achieves accuracy comparable to the baseline methods. On medium-difficulty tasks such as MATH500 and AMC2023, its accuracy is essentially on par with SC, with only minor fluctuations (within ±1 percentage point). On the more challenging datasets AIME2024/2025 and GPQA-Diamond, SeerSC also sustains stable performance without noticeable degradation. Given the stochasticity introduced by high-temperature decoding, the overall accuracy level remains comparable to that of the baselines.

\paragraph{SeerSC reduces token usage.}
SeerSC consistently surpasses SC in terms of token efficiency. By adaptively determining the number of samples in advance, it avoids a large amount of unnecessary generation. As a result, SeerSC can substantially reduce token usage while maintaining accuracy. On MATH500, AMC2023, and GPQA-D across all models, it achieves lower token consumption than all other baselines. On AIME2024/2025, although the token usage is slightly higher than that of AC and ESC, it remains considerably lower than SC.

\paragraph{SeerSC significantly lowers latency.}
The core advantage of SeerSC lies in its inference efficiency. Compared to SC, SeerSC substantially reduces latency across the majority of models and datasets. For example, on DeepSeek-R1-Distill-Qwen-7B, SeerSC achieves an average latency of only 5.3s on MATH500, a clear improvement over 8.4s of SC, representing a 37\% reduction. On Qwen3-4B (non-thinking), the latency advantage is even more pronounced: SeerSC requires only 3.7s on GPQA-D, compared to 6.5s of SC, yielding a reduction of approximately 43\%.By contrast, AC and ESC suffer from severe latency bottlenecks due to their sequential generation design, consistently running slower than SC across models and datasets. On Qwen3-4B (non-thinking), both methods take on average twice as long as SC. On DeepSeek-R1-Distill-Llama-8B and Qwen3-4B (thinking) for AIME2024/2025 and GPQA-D, where long reasoning trajectories dominate computation, the latencies of AC and ESC become comparable to SC. Nevertheless, under these settings, SeerSC still achieves significantly faster inference than all baseline methods.

\paragraph{Overall.}
SeerSC achieves accuracy comparable to SC, AC, and ESC, while significantly outperforming all baselines in terms of token and latency efficiency. It thus provides a more practical and effective trade-off for applications where both low latency and high accuracy are critical.

\section{Analysis}
\subsection{Latency Analysis}

\paragraph{The latency of System 1 is negligible.}

\begin{figure}[h]
    \centering
    \includegraphics[width=0.98\linewidth]{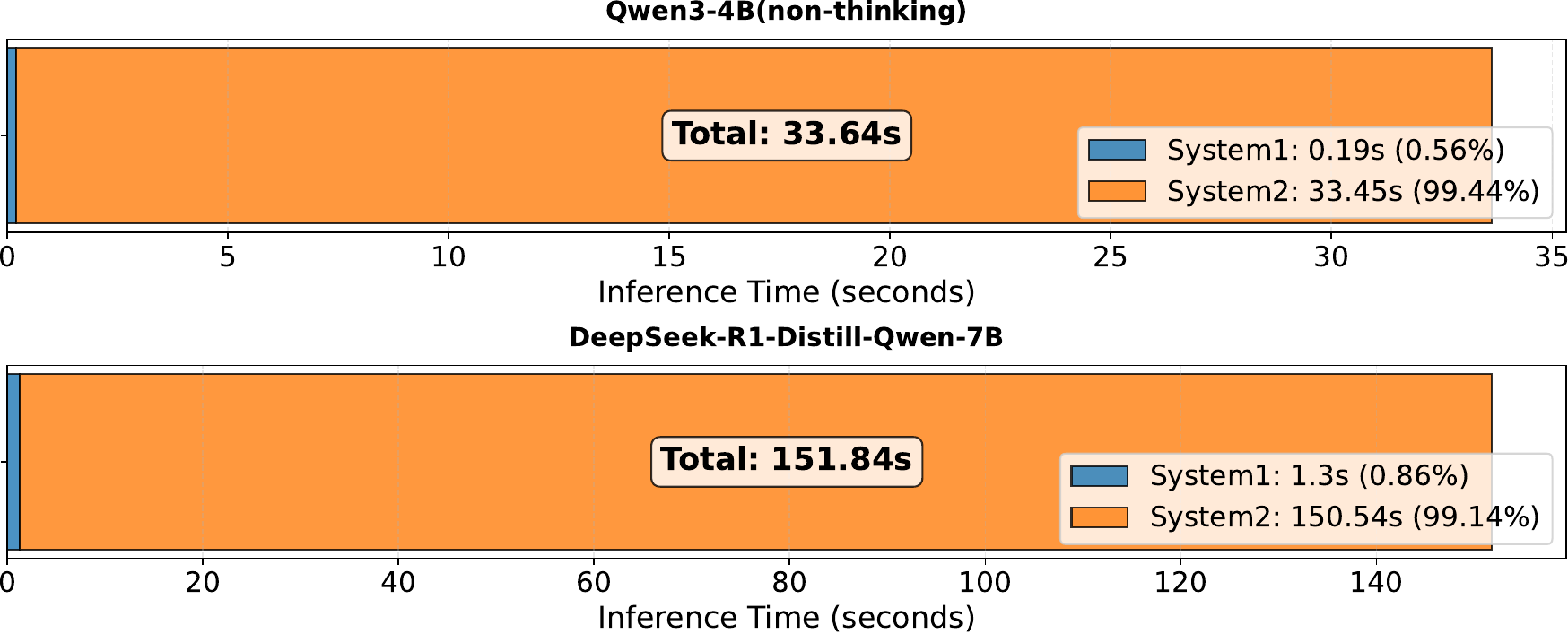}
    \caption{Latency comparison of System~1 and System~2 in SeerSC on AIME-2024, using DeepSeek-R1-Distill-Qwen-7B and Qwen3-4B (non-thinking).}
    \label{fig:System_1_latency}
\end{figure}

As shown in Figure~\ref{fig:System_1_latency}, the latency of System 1 accounts for less than 1\% of the total latency of SeerSC. In practical applications, the overhead of System 1 would be even smaller, since the KV cache produced during its prefilling stage can be reused in System 2. As a result, the additional delay corresponds only to decoding a few tokens and is therefore negligible (see Appendix~\ref{sec:ttft_analysis} for a evaluation of Time to First Token).

\paragraph{SeerSC exhibits better scaling on the latency–accuracy trade-off}

\begin{figure*}[h]
    \centering
    \includegraphics[width=\textwidth]{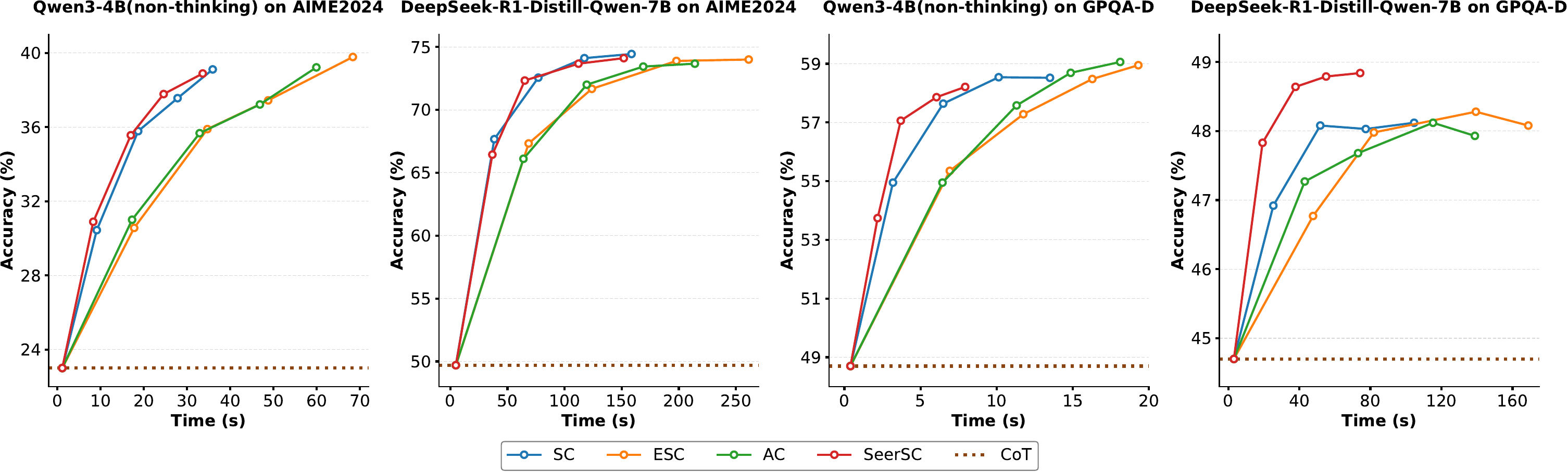}
    \caption{Latency scaling results of SeerSC compared with baseline methods on AIME-2024 and GPQA-D using Qwen3-4B (non-thinking) and DeepSeek-R1-Distill-Qwen-7B. SeerSC matches or exceeds SC under low-latency conditions, reduces latency by about 50\% compared to AC and ESC on AIME-2024, and achieves the same accuracy with up to 75\% lower latency on GPQA-D.}
    \label{fig:latency_scaling}
\end{figure*}

We conduct latency scaling experiments on Qwen3-4B (non-thinking) and DeepSeek-R1-Distill-Qwen-7B using  AIME-2024 and GPQA-D, evaluating SeerSC against baseline methods. As illustrated in Figure \ref{fig:latency_scaling}, the CoT method serves as the starting point of extended performance, where allocating more computation time to the model enables us to examine the scaling behavior of each method.

The results show that on AIME2024, a challenging mathematical reasoning benchmark, SeerSC exhibits scaling performance comparable to SC, while achieving slightly higher accuracy under low-latency conditions. 
Moreover, both SeerSC and SC reduce latency by about 50\% compared to AC and ESC on this task. 
On GPQA-D, a graduate-level scientific reasoning task, SeerSC demonstrates superior scaling performance. Specifically, on DeepSeek-R1-Distill-Qwen-7B, SeerSC reaches the convergence accuracy of SC with only 50\% of the latency. 
Compared to AC and ESC, SeerSC achieves the same accuracy with a 75\% reduction in latency.

In summary, across different models and datasets, SeerSC consistently exhibits favorable latency scaling properties, significantly outperforming the sequential methods AC and ESC.

\subsection{Ablation Study}

\paragraph{Temperature in System 1 can balance accuracy and latency.}

\begin{figure}[h]
    \centering
    \includegraphics[width=0.5\textwidth]{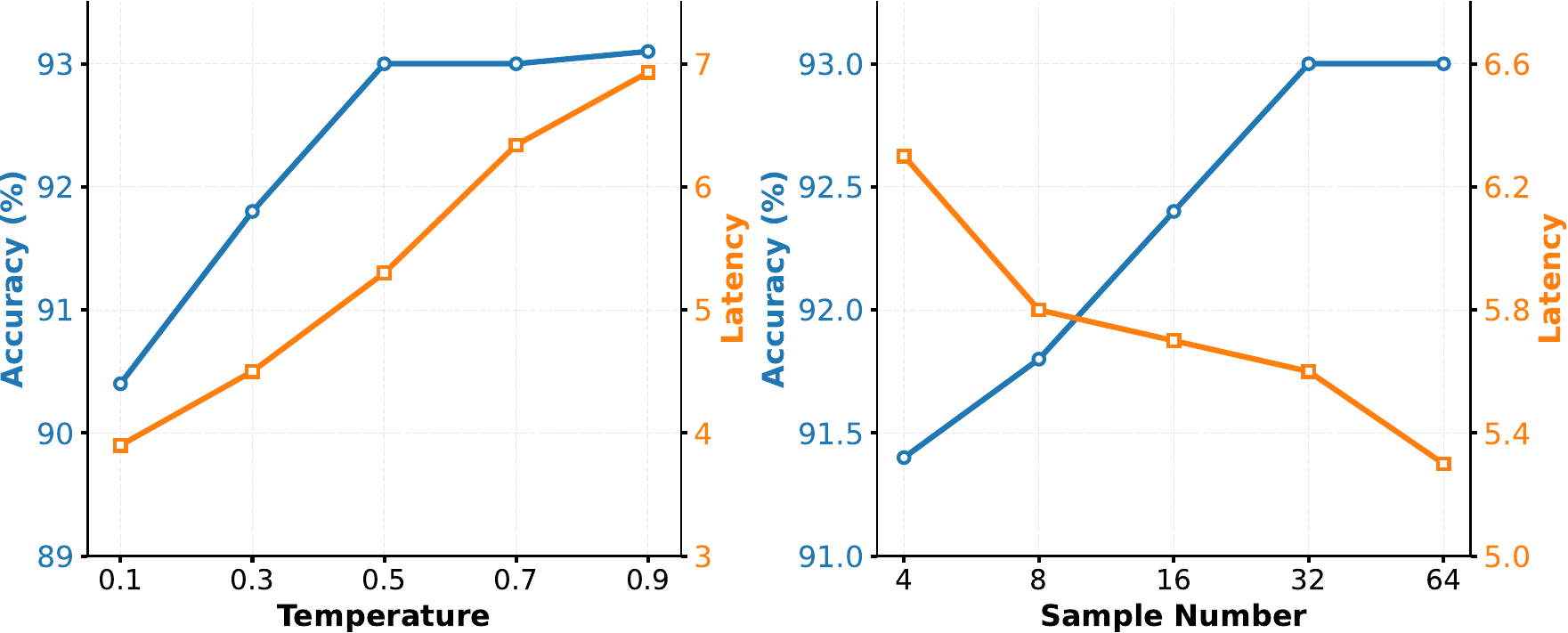}
    \caption{Ablation study on System~1 using DeepSeek-R1-Distill-Qwen-7B on MATH500. (Left) effect of sampling temperature; (Right) effect of sampling size.}
    \label{fig:ablation}
\end{figure}

 The sampling temperature used in System1 primarily affects the distribution of anticipatory entropy, which in turn influences the load allocation of System 2. As illustrated in Figure \ref{fig:ablation}(left), a low temperature shifts the entropy distribution toward lower values, reducing latency but also lowering accuracy. In contrast, a high temperature shifts the distribution toward higher values, increasing latency while improving accuracy. Therefore, selecting an appropriate temperature is crucial for achieving a favorable trade-off.

\paragraph{Larger sampling size in System 1 leads to more accurate entropy estimation.}

According to the Monte Carlo sampling principle, increasing the number of samples allows the estimated entropy distribution to approach its expected value, thereby improving stability. As shown in Figure \ref{fig:ablation}(right), with more samples, accuracy gradually increases while latency decreases. However, once an appropriate sampling size is reached, further increasing it provides little benefit and instead adds extra latency to System 1.

\paragraph{Incorporating confidence-weighted entropy estimation improves performance.}

\begin{table}[t]
\centering
\small
\caption{Effect of confidence-weighted entropy estimation in System~1 using DeepSeek-R1-Distill-Qwen-7B on MATH500 and AIME-2024.}
\label{tab:conf_weight}
\begin{tabularx}{\linewidth}{lXXXX}

\toprule
\multirow{2}{*}{Method} & \multicolumn{2}{c}{\textbf{MATH500}} & \multicolumn{2}{c}{\textbf{AIME2024}} \\
\cmidrule(lr){2-3} \cmidrule(lr){4-5}
 & \textit{Acc.}$\uparrow$ & \textit{Lat.}$\downarrow$ & \textit{Acc.}$\uparrow$ & \textit{Lat.}$\downarrow$ \\
\midrule
w.\ Confidence Weighted   & \textbf{93.0} & \textbf{5.3} & \textbf{74.11} & \textbf{151.8} \\
w.o.\ Confidence Weighted  & 92.2 & 5.6 & 73.56 & 156.5  \\
\bottomrule
\end{tabularx}
\end{table}

 The Shannon entropy distribution in System 1 is constrained by the number of answer categories, causing entropy values to accumulate around discrete points, which is suboptimal for load allocation in System 2. By integrating confidence weighting, the entropy distribution becomes smoother and more continuous, while also increasing the contribution of high-confidence answers in the entropy calculation. As shown in Table \ref{tab:conf_weight}, downstream task results indicate that confidence-weighted entropy estimation not only improves performance but also moderately reduces latency.

\subsection{Orthogonality}
\label{sec:orthogonality}

\begin{table*}[t]
\scriptsize
\centering
\caption{Orthogonality study of SeerSC with weighted vote and path pruning on DeepSeek-R1-Distill-Qwen-7B and Qwen3-4B (non-thinking). 
We report accuracy (\%), average tokens ($\times 10^3$), and average latency (s) on five datasets. 
Results show that SeerSC can be effectively combined with these strategies, improving accuracy or efficiency without introducing additional latency.}
\label{tab:Orthogonality}
\begin{tabularx}{\textwidth}{c*{15}{>{\centering\arraybackslash}X}}

\toprule
\multirow{2}{*}{\textbf{Methods}}
& \multicolumn{3}{c}{\textbf{MATH500}} 
& \multicolumn{3}{c}{\textbf{AIME2024}} 
& \multicolumn{3}{c}{\textbf{AIME2025}} 
& \multicolumn{3}{c}{\textbf{AMC2023}} 
& \multicolumn{3}{c}{\textbf{GPQA-D}} \\
\cmidrule(lr){2-4}\cmidrule(lr){5-7}\cmidrule(lr){8-10}\cmidrule(lr){11-13}\cmidrule(lr){14-16}
& \textit{Acc.}$\uparrow$ & \textit{Token}$\downarrow$ & \textit{Lat.$\downarrow$}
& \textit{Acc.}$\uparrow$ & \textit{Token}$\downarrow$ & \textit{Lat.$\downarrow$}
& \textit{Acc.}$\uparrow$ & \textit{Token}$\downarrow$ & \textit{Lat.$\downarrow$}
& \textit{Acc.}$\uparrow$ & \textit{Token}$\downarrow$ & \textit{Lat.$\downarrow$}
& \textit{Acc.}$\uparrow$ & \textit{Token}$\downarrow$ & \textit{Lat.$\downarrow$}\\
\midrule

\rowcolor{gray!10}
\multicolumn{16}{c}{\textit{DeepSeek-R1-Distill-Qwen-7B}} \\
\midrule
SeerSC & 93.0 & 17.7 & \textbf{5.3}  & 74.1 & 315.6 & 151.8 & 60.2 & 356.0 & \textbf{173.8} & 91.5 & 33.5 & \textbf{9.4}  & 48.6 & 95.3 & \textbf{37.8} \\
+ Weighted Vote & \textbf{93.8} & 17.8 & \textbf{5.3} & \textbf{75.7} & 315.8 & 152.0 & \textbf{60.5} & 357.1 & 174.0 & \textbf{92.2} & \textbf{33.4} & \textbf{9.4} & \textbf{49.0} & 95.7 & 38.1 \\
+ Path Pruning & 93.4 & \textbf{17.5} & \textbf{5.3} & 74.2 & \textbf{306.1} & \textbf{151.7} & 58.3 & \textbf{338.9} & \textbf{173.8} & 91.5 & \textbf{33.4} & \textbf{9.4} & 48.2 & \textbf{91.1} & \textbf{37.8} \\
\midrule

\rowcolor{gray!10}
\multicolumn{16}{c}{\textit{Qwen3-4B (non-thinking)}} \\
\midrule
SeerSC & 88.1 & \textbf{4.4}  & 1.5 & 38.9 & 93.6  & 33.6 & 22.9 & 47.8  & \textbf{16.1} & 76.3 & 9.5  & \textbf{3.3} & 57.1 & 13.0 & \textbf{3.7} \\
+ Weighted Vote & \textbf{88.8} & 4.5 & \textbf{1.4} & \textbf{41.9} & 93.7 & \textbf{33.4} & \textbf{23.9} & 47.8 & 16.2 & \textbf{77.9} & 9.6 & \textbf{3.3} & \textbf{57.5} & 13.2 & 3.8 \\
+ Path Pruning & 88.2 & \textbf{4.4} & 1.5 & 38.9 & \textbf{87.1} & 33.9 & 22.7 & \textbf{44.8} & \textbf{16.1} & 76.8 & \textbf{9.4} & \textbf{3.3} & 56.5 & \textbf{11.8} & 3.9 \\
\bottomrule
\end{tabularx}
\end{table*}

It is worth noting that SeerSC is orthogonal to several existing parallel test-time scaling optimization techniques, such as weighted voting and path pruning. We conducted experiments to validate this claim. For extended experiments regarding orthogonality, please refer to Appendix~\ref{sec:baseline_weighted} and \ref{sec:compatibility}.

\paragraph{SeerSC is orthogonal to weighted vote.}
Existing weighted voting assigns confidence-based weights to paths, allowing high-confidence samples to exert greater influence and improve accuracy. \citep{wan2025reasoning, zhou2025bridging}. Building on standard SeerSC, we incorporated the weighted voting strategy from \citep{fu2025deep}, which leverages tail confidence—the confidence scores of the last window in a path—to determine its weight in voting. As shown in Table \ref{tab:Orthogonality}, both DeepSeek-R1-Distill-Qwen-7B and Qwen3-4B (non-thinking) achieved accuracy improvements across datasets. Notably, on AIME-2024, Qwen3-4B (non-thinking) outperformed vanilla SeerSC by +3 percentage points. Moreover, combining the two methods does not affect token consumption or latency efficiency.

\paragraph{SeerSC is orthogonal to path pruning.}
Existing path pruning approaches generally prune low-quality reasoning paths during parallel generation in order to reduce token usage and improve voting accuracy \citep{wang2025sampling, hong2025slim}. On top of standard SeerSC, we adopted the pruning strategy from \citep{fu2025deep}, which measures the minimum confidence along a path. If this value falls below a fixed threshold, the path is pruned. Specifically, the pruning threshold is set to 7 for DeepSeek-R1-Distill-Qwen-7B and 10.5 for Qwen3-4B (non-thinking). As shown in Table \ref{tab:Orthogonality}, both models achieved improved token efficiency on AIME-2024/2025 and GPQA-D, and in most tasks, a slight improvement in voting accuracy was also observed.

\section{Related Work}
\subsection{Test-Time Scaling}
Test-Time Scaling (TTS) allocates extra computation at inference to improve performance~\citep{snell2024scaling}, 
without parameter updates, and can be broadly categorized into \emph{sequential} and \emph{parallel} methods~\citep{zhang2025survey}.

Sequential TTS refines a single candidate solution step by step. 
Chain-of-Thought~\citep{wei2022chain} is the most representative approach, 
where intermediate reasoning steps are explicitly generated to improve accuracy and interpretability. 
Extensions such as Self-Refine~\citep{madaan2023self} further iteratively improve answers, 
though excessive reasoning can sometimes hurt performance~\citep{chen2024not}.  

Parallel TTS generates multiple candidate solutions in parallel and aggregates them to form the final answer. 
Self-Consistency~\citep{wang2022self} selects the majority outcome among sampled paths, 
while Best-of-$N$ (BoN)~\citep{lightman2023let} uses a reward model to select the best candidate. 
These methods improve robustness but often face diminishing returns as the sample size increases~\citep{snell2024scaling}.

\subsection{Efficient Optimization of Parallel TTS.} 
Recent work has focused on improving the efficiency of TTS~\citep{zeyularge, xu2025adaptive}. 
Existing approaches to optimizing parallel TTS can be broadly grouped into three categories:

\textbf{(1) \emph{Adaptive sampling.}}
These methods dynamically adjust the number of samples $N$. 
For instance, AC \citep{aggarwal2023let} and ESC \citep{liescape} stop sampling early once a consensus emerges among recent answers; 
DSC~\citep{wang2024make} uses prompting to decide the sample size;
RASC \citep{wan2025reasoning} trains a classifier to decide early stopping. 
While these sequential sampling methods hinders GPU parallelization 
and introduces additional latency \citep{hong2025slim}.  

\textbf{(2) \emph{Path pruning.}}
Another line of work prunes low-quality reasoning paths during generation to reduce tokens 
and improve aggregation accuracy. 
For example, Fast BoN \citep{sun2024fast} employs a reward model to retain promising trajectories, 
ST-BoN \citep{wang2025sampling} measures pairwise distances between sampled paths to prune inconsistent ones, 
and Slim-SC \citep{hong2025slim} prunes similar or redundant trajectories. 

\textbf{(3) \emph{Hybrid methods.}}
Some methods combine adaptive sampling and pruning. 
For instance, RPC \citep{zhou2025bridging} uses perplexity-based pruning with dynamic sampling; 
DeepThink \citep{fu2025deep} incorporates multiple confidence metrics to prune low-confidence paths, 
adaptively decides whether to stop sampling, and applies weighted voting to improve accuracy.  


\section{Conclusion}
We proposed \textbf{SeerSC}, a dynamic self-consistency framework. 
Compared with sequential adaptive methods, SeerSC achieves lower latency without sacrificing accuracy, and it remains orthogonal and complementary to confidence-weighted voting and path pruning, enabling further performance gains.


\section*{Limitations}
Although the proposed method demonstrates advantages in both token consumption and latency,
there remains scope for further improvement.
In domains with verifiable outcomes (e.g., mathematics), the answer entropy from System 1 is directly accessible.
However, in broader domains like code generation, existing entropy-based dynamic TTS methods do not provide reliable estimation.
Our future work will focus on extending the application of dynamic test-time scaling to a wider range of scenarios by leveraging techniques such as clustering and semantic vectors.


\bibliography{custom}

\appendix

\begin{table*}[h]
\centering
\small
\caption{Optimal $N$ for SC on different models and datasets.}
\label{tab:optimalN}
\begin{tabularx}{\textwidth}{l*{5}{>{\centering\arraybackslash}X}}
\toprule
Model & MATH500 & AIME2024 & AIME2025 & AMC2023 & GPQA-D \\
\midrule
DeepSeek-R1-Distill-Qwen-7B   & 8  & 32 & 32 & 8  & 16 \\
DeepSeek-R1-Distill-Llama-8B  & 8  & 32 & 32 & 8  & 8  \\
Qwen3-4B (thinking)           & 8  & 16 & 16 & 8  & 8  \\
Qwen3-4B (non-thinking)       & 8  & 32 & 16 & 8  & 16 \\
\bottomrule
\end{tabularx}
\end{table*}

\begin{table*}[t]
    \centering
    \small
    \caption{Hyperparameter Sensitivity Analysis on the MATH training dataset using DeepSeek-R1-Distill-Qwen-7B. 
    We identify the optimal configuration step-by-step: first determining $\tau_1$, then tuning $\tau_2$.}
    \label{tab:sensitivity_analysis}

    \begin{subtable}{\linewidth}
        \centering
        \caption{Impact of the First Threshold $\tau_1$ (single threshold setting). 
        $\tau_1 = 1/10$ offers a favorable balance between SC and CoT behaviors.}
        \label{tab:threshold_single}
        \begin{tabular}{lcccccc}
        \toprule
        \textbf{Metric} & \textbf{(0, -)} & \textbf{(1/20, -)} & \textbf{(1/10, -)} & \textbf{(1/5, -)} & \textbf{(1/3, -)} & \textbf{(1, -)} \\
        \midrule
        Acc. (\%) & 93.9 & 93.8 & \textbf{93.3} & 92.8 & 92.0 & 90.0 \\
        Lat. (s) & 8.4 & 7.8 & \textbf{7.4} & 5.8 & 4.2 & 1.1 \\
        \bottomrule
        \end{tabular}
    \end{subtable}

    \bigskip

    \begin{subtable}{\linewidth}
        \centering
        \caption{Impact of the Second Threshold $\tau_2$ (with $\tau_1$ fixed at $1/10$). 
        The configuration $(1/10, 1/3)$ further reduces latency with negligible accuracy loss.}
        \label{tab:threshold_dual}
        \begin{tabular}{lcccccc}
        \toprule
        \textbf{Metric} & \textbf{(1/10, -)} & \textbf{(1/10, 1/6)} & \textbf{(1/10, 1/5)} & \textbf{(1/10, 1/3)} & \textbf{(1/10, 1/2)} & \textbf{(1/10, 1)} \\
        \midrule
        Acc. (\%) & 93.3 & 93.2 & 93.0 & \textbf{93.0} & 92.4 & 92.0 \\
        Lat. (s) & 7.4 & 6.6 & 6.1 & \textbf{5.3} & 4.3 & 3.6 \\
        \bottomrule
        \end{tabular}
    \end{subtable}
\end{table*}

\section{Implementation Details}
\label{sec:details}

\subsection{Optimal $N$ for SC}
Table~\ref{tab:optimalN} summarizes the optimal sampling number $N$ for the SC across models and datasets.

\subsection{Implementation details of SeerSC}
For DeepSeek-R1-Distill-Qwen-7B and DeepSeek-R1-Distill-Llama-8B, 
System 1 is run with a generation temperature of \textbf{$0.5$}, 
while for Qwen3-4B (thinking) and Qwen3-4B (non-thinking), 
System 1 is run with the temperature of \textbf{$1.0$}. 
For all models and datasets, the number of direct answers in System~1 is fixed at \textbf{$64$}.

\subsection{System 1 Implementation for Reasoning Models}
\label{sec:system1_implementation}

Reasoning models (e.g., DeepSeek-R1 series or QwQ) typically generate an extended reasoning process enclosed within \texttt{<think>} and \texttt{</think>} tags before producing the final answer. However, System 1 requires rapid, direct answer generation to estimate entropy efficiently.

To enable this, we employ a forced decoding strategy. Specifically, we append a sequence to the model input that simulates the completion of the thinking process and prompts the immediate start of the final answer. The appended sequence is:

\begin{quote}
    \small
    \texttt{\textbf{\textbackslash n</think>So the final answer is \textbackslash}}
\end{quote}

By forcing the model to process this sequence immediately after the user query, we bypass the generation of the reasoning trace. The model proceeds as if it has already completed its reasoning, directly generating the content within \texttt{\textbackslash boxed\{\}}.

\paragraph{Example.}
Given the input query:
\begin{quote}
    \small
    \texttt{<|User|>How many positive... within \textbackslash boxed\{\}. <|Assistant|><think>}
\end{quote}
A standard generation would produce:
\begin{quote}
    \small
    \texttt{(thinking...)</think>So the final answer is \textbackslash boxed\{1\}.}
\end{quote}
With our suffix injection, the effective input becomes:
\begin{quote}
    \small
    \texttt{<|User|>...<|Assistant|><think>\textbf{\textbackslash n</think>So the final answer is \textbackslash}}
\end{quote}
And the model output becomes:
\begin{quote}
    \small
    \texttt{\textbf{\textbackslash boxed\{1\}.}}
\end{quote}
This technique ensures that System 1 inference remains computationally lightweight and fast.

\begin{table*}[h]
\scriptsize
\centering
\caption{Experimental results on \textbf{QwQ-32B} model. 
SeerSC demonstrates strong generalizability, achieving significant reductions in token usage and latency on diverse datasets, while maintaining accuracy comparable to baselines.
We report accuracy (\%), average tokens ($\times 10^3$), and average latency (s).}
\label{tab:qwq_results}
\begin{tabularx}{\textwidth}{c*{15}{>{\centering\arraybackslash}X}}

\toprule
\multirow{2}{*}{\textbf{Methods}}
& \multicolumn{3}{c}{\textbf{MATH500}} 
& \multicolumn{3}{c}{\textbf{AIME2024}} 
& \multicolumn{3}{c}{\textbf{AIME2025}} 
& \multicolumn{3}{c}{\textbf{AMC2023}} 
& \multicolumn{3}{c}{\textbf{GPQA-D}} \\
\cmidrule(lr){2-4} \cmidrule(lr){5-7} \cmidrule(lr){8-10} \cmidrule(lr){11-13} \cmidrule(lr){14-16}
& \textit{Acc.}$\uparrow$ & \textit{Token}$\downarrow$ & \textit{Lat.$\downarrow$}
& \textit{Acc.}$\uparrow$ & \textit{Token}$\downarrow$ & \textit{Lat.$\downarrow$}
& \textit{Acc.}$\uparrow$ & \textit{Token}$\downarrow$ & \textit{Lat.$\downarrow$}
& \textit{Acc.}$\uparrow$ & \textit{Token}$\downarrow$ & \textit{Lat.$\downarrow$}
& \textit{Acc.}$\uparrow$ & \textit{Token}$\downarrow$ & \textit{Lat.$\downarrow$} \\
\midrule

\rowcolor{gray!10}
\multicolumn{16}{c}{\textit{QwQ-32B}} \\
\midrule
CoT & 91.1 & -- & -- & 65.1 & -- & -- & 51.8 & -- & -- & 81.1 & -- & -- & 63.5 & -- & -- \\
\hdashline
SC  & \textbf{93.6} & 33.3 & 16.2 & 74.0 & 93.4 & 95.3 & 66.0 & 101.3 & 87.6 & 86.0 & 51.7 & 29.6 & \textbf{66.4} & 69.6 & 58.0 \\
AC  & 93.5 & 21.0 & 19.1 & 73.3 & \textbf{73.6} & 135.2 & 65.3 & \textbf{91.4} & 123.7 & 86.0 & 36.9 & 41.2 & 66.1 & \textbf{53.9} & 70.5 \\
ESC & 93.4 & 23.9 & 22.2 & 73.6 & 78.5 & 127.1 & 65.1 & 93.8 & 115.5 & 86.5 & 40.4 & 45.8 & 66.1 & 58.6 & 77.9 \\
\rowcolor{blue!10}
SeerSC & 93.0 & \textbf{19.5} & \textbf{9.6}  & \textbf{74.4} & 88.3 & \textbf{90.6} & \textbf{66.2} & 95.4 & \textbf{83.2} & 85.5 & \textbf{36.7} & \textbf{19.1} & 66.3 & 59.7 & \textbf{43.4} \\
\bottomrule
\end{tabularx}
\end{table*}

\section{Hyperparameter Sensitivity Analysis}
\label{sec:sensitivity}

The entropy thresholds $\tau_1$ and $\tau_2$ in Equation~\ref{eq:budget} are critical hyperparameters that determine the trade-off between accuracy and latency. We empirically determined these values based on performance on the MATH training dataset using the DeepSeek-R1-Distill-Qwen-7B model.

\paragraph{Impact of the First Threshold ($\tau_1$).}
We first investigated the impact of a single threshold $\tau_1$ (effectively setting $\tau_2 = \tau_1$, resulting in a binary budget allocation of either 1 or $N$). As shown in Table~\ref{tab:threshold_single}, increasing $\tau_1$ leads to more samples being assigned a budget of 1. Consequently, the method transitions from the behavior of SC (high accuracy, high latency) to that of CoT (low accuracy, low latency). A threshold of $1/10$ provides a strong initial filtering, significantly reducing latency while preserving most of the accuracy gains.

\paragraph{Impact of the Second Threshold ($\tau_2$).}
To achieve a finer-grained budget allocation, we introduced the second threshold $\tau_2$ (assigning a budget of $N/2$ for samples with entropy between $\tau_1$ and $\tau_2$). We fixed $\tau_1 = 1/10$ and varied $\tau_2$. As presented in Table~\ref{tab:threshold_dual}, introducing $\tau_2$ further optimizes the efficiency. The setting $(\tau_1=1/10, \tau_2=1/3)$ achieves a significant reduction in latency (from 7.4s to 5.3s) compared to using a single threshold, with only a negligible drop in accuracy (0.3\%).

\paragraph{Summary.}
These empirical results justify our choice of $\tau_1 = \frac{1}{10}\log M$ and $\tau_2 = \frac{1}{3}\log M$. While this configuration may not be the global optimum for every specific scenario, our extensive experiments across diverse datasets and model scales (ranging from 4B to 32B) demonstrate its consistent effectiveness and robustness. We adopt these fixed values to avoid the additional computational overhead associated with per-task hyperparameter searching on validation sets. The exploration of more fine-grained or adaptive thresholding mechanisms remains a promising avenue for our future work.

\section{Additional Experiments}
\label{sec:additional_experiments}

\subsection{Generalizability to Larger Models}
\label{sec:large_models}

To assess generalizability, we extended our evaluation to the larger-scale \textbf{QwQ-32B} model with a sampling budget of $N=8$. As shown in Table~\ref{tab:qwq_results}, the core conclusions observed in smaller models remain valid: SeerSC maintains accuracy parity with the SC baseline while significantly reducing computational costs. This confirms that SeerSC's adaptive budget allocation remains robust and effective at larger model scales.

\subsection{Impact on Time to First Token (TTFT)}
\label{sec:ttft_analysis}

We further investigated the impact of the System 1 pre-estimation stage on the Time to First Token (TTFT), a critical metric for streaming applications. We measured the TTFT on a single AIME2024 query using \textit{DeepSeek-R1-Distill-Qwen-7B} served by \texttt{vLLM}.

As shown in Table~\ref{tab:ttft_comparison}, although SeerSC requires a pre-estimation step (generating 64 direct answers), the increase in TTFT is minimal. 
Specifically, the TTFT increases from 0.022s (SC) to 0.234s (SeerSC), adding an overhead of only $\approx 0.2$s. 
Given that the average completion time for a complex reasoning task (e.g., AIME) is typically around 30 seconds, this slight delay is negligible relative to the total inference latency and does not perceptibly affect the user experience.

\begin{table}[h]
    \centering
    \small
    \caption{Comparison of Time to First Token (TTFT) between SC and SeerSC on a single AIME query. The overhead introduced by System 1 is negligible ($\approx 0.2$s).}
    \label{tab:ttft_comparison}
    \begin{tabular}{lcc}
        \toprule
        \textbf{Method} & \textbf{TTFT (s)} & \textbf{Overhead (s)} \\
        \midrule
        SC & 0.022 & -- \\
        SeerSC & 0.234 & +0.212 \\
        \bottomrule
    \end{tabular}
\end{table}

\begin{table*}[t]
\scriptsize
\centering
\caption{Impact of Weighted Voting on baselines (DeepSeek-R1-Distill-Qwen-7B). Values in parentheses denote changes relative to vanilla versions. Despite accuracy gains in baselines, SeerSC retains superior latency efficiency.}
\label{tab:baseline_weighted_results}
\begin{tabularx}{\textwidth}{c*{6}{>{\centering\arraybackslash}X}}
\toprule
\multirow{2}{*}{\textbf{Methods}}
& \multicolumn{2}{c}{\textbf{MATH500}} 
& \multicolumn{2}{c}{\textbf{AMC2023}} 
& \multicolumn{2}{c}{\textbf{GPQA-D}} \\
\cmidrule(lr){2-3} \cmidrule(lr){4-5} \cmidrule(lr){6-7}
& \textit{Acc. (\%)}$\uparrow$ & \textit{Lat. (s)}$\downarrow$
& \textit{Acc. (\%)}$\uparrow$ & \textit{Lat. (s)}$\downarrow$
& \textit{Acc. (\%)}$\uparrow$ & \textit{Lat. (s)}$\downarrow$ \\
\midrule

SC & 93.9 & 8.4 & 91.6 & 13.2 & 48.1 & 51.8 \\
+ Weighted Vote & 94.1 \scriptsize{(+0.2)} & 8.4 & \textbf{92.5} \scriptsize{(+0.9)} & 13.2 & 48.7 \scriptsize{(+0.6)} & 52.1 \\
\midrule
AC & 93.8 & 12.4 & 91.3 & 24.8 & 47.7 & 73.1 \\
+ Weighted Vote & \textbf{94.0} \scriptsize{(+0.2)} & 12.5 & 92.3 \scriptsize{(+1.0)} & 25.2 & 47.7 \scriptsize{(+0.0)} & 73.5 \\
\midrule
ESC & 93.6 & 13.9 & 91.1 & 26.7 & 48.0 & 82.0 \\
+ Weighted Vote & 93.4 \scriptsize{(-0.2)} & 13.8 & 92.1 \scriptsize{(+1.0)} & 26.7 & 48.6 \scriptsize{(+0.6)} & 82.3 \\
\midrule
\rowcolor{blue!10}
SeerSC & 93.0 & \textbf{5.3} & 91.5 & \textbf{9.4} & 48.6 & \textbf{37.8} \\
\rowcolor{blue!10}
+ Weighted Vote & 93.8 \scriptsize{(+0.8)} & \textbf{5.3} & 92.2 \scriptsize{(+0.7)} & \textbf{9.4} & \textbf{49.0} \scriptsize{(+0.4)} & 38.1 \\

\bottomrule
\end{tabularx}
\end{table*}

\begin{table*}[t]
    \centering
    \small
    \caption{Experimental results of combining SeerSC with other TTS paradigms. 
    The results demonstrate that SeerSC is compatible with both reward model-based and length-controlled methods.}
    \label{tab:bon_tokenskip}

    \begin{subtable}{\linewidth}
        \centering
        \caption{Combination with \textbf{Best-of-N} (Generator: DeepSeek-R1-Distill-Qwen-7B; RM: Skywork-Reward-V2-Qwen3-8B). 
        BoN improves accuracy, at the cost of scoring latency.}
        \begin{tabularx}{\linewidth}{l *{6}{>{\centering\arraybackslash}X}}
            \toprule
            \multirow{2}{*}{\textbf{Method}} & \multicolumn{3}{c}{\textbf{MATH500}} & \multicolumn{3}{c}{\textbf{GPQA-D}} \\
            \cmidrule(lr){2-4} \cmidrule(lr){5-7}
             & \textit{Acc.} (\%) & \textit{Token} ($10^3$) & \textit{Lat.} (s) & \textit{Acc.} (\%) & \textit{Token} ($10^3$) & \textit{Lat.} (s) \\
            \midrule
            SeerSC & 93.0 & 17.7 & \textbf{5.3} & 48.6 & 95.3 & \textbf{37.8} \\
            + Best-of-N & \textbf{93.4} \scriptsize{(+0.4)} & 17.7 & 7.3 & \textbf{51.0} \scriptsize{(+2.4)} & 95.3 & 48.7 \\
            \bottomrule
        \end{tabularx}
    \end{subtable}

    \bigskip 

    \begin{subtable}{\linewidth}
        \centering
        \caption{Combination with \textbf{TokenSkip} (Model: Qwen2.5-7B-Instruct fine-tuned with TokenSkip). 
        TokenSkip reduces token usage significantly, demonstrating orthogonality with SeerSC.}
        \begin{tabularx}{\linewidth}{l *{6}{>{\centering\arraybackslash}X}}
            \toprule
            \multirow{2}{*}{\textbf{Method}} & \multicolumn{3}{c}{\textbf{GSM8K}} & \multicolumn{3}{c}{\textbf{MATH500}} \\
            \cmidrule(lr){2-4} \cmidrule(lr){5-7}
             & \textit{Acc.} (\%) & \textit{Token} & \textit{Lat.} (s) & \textit{Acc.} (\%) & \textit{Token} & \textit{Lat.} (s) \\
            \midrule
            SeerSC & 92.9 & 1437 & 0.8 & \textbf{78.6} & 2726 & 0.5 \\
            + TokenSkip & 91.0 \scriptsize{(-1.9)} & \textbf{1205} \scriptsize{(-16\%)} & 0.8 & 76.0 \scriptsize{(-2.6)} & \textbf{2379} \scriptsize{(-13\%)} & 0.5 \\
            \bottomrule
        \end{tabularx}
    \end{subtable}
\end{table*}

\subsection{Impact of Weighted Voting on Baselines}
\label{sec:baseline_weighted}

We extend the confidence-based weighted voting \citep{fu2025deep} to baselines (SC, AC, ESC) on DeepSeek-R1-Distill-Qwen-7B to evaluate potential gains. 
As shown in Table~\ref{tab:baseline_weighted_results}, while weighted voting yields marginal accuracy improvements for baselines, it fails to address their inherent latency bottlenecks. 
In contrast, \textbf{SeerSC maintains a significant efficiency advantage}—e.g., on GPQA-D, it is $1.4\times$ faster than SC and $\approx 2\times$ faster than AC/ESC—confirming that our efficiency gains stem from proactive budget allocation rather than the voting mechanism.

\subsection{Compatibility with Other TTS Paradigms}
\label{sec:compatibility}

In Section~\ref{sec:orthogonality}, we demonstrated that SeerSC is orthogonal to parallel optimization techniques such as weighted voting and path pruning. Beyond these strategies, SeerSC serves as a flexible budget allocation framework that can also be integrated with broader Test-Time Scaling (TTS) paradigms. To demonstrate this extensibility, we conducted experiments combining SeerSC with Reward Model-based selection (Best-of-N) and Length-Controlled scaling (TokenSkip).

\paragraph{Combining with Best-of-N.}
We integrated SeerSC with Best-of-N (BoN) using DeepSeek-R1-Distill-Qwen-7B as the generator and Skywork-Reward-V2-Qwen3-8B \citep{liu2025skywork} as the reward model. Instead of majority voting, we used the reward model to rank the candidates generated within the SeerSC budget.

As shown in Table~\ref{tab:bon_tokenskip}(a), this combination yields consistent accuracy improvements, particularly on the challenging GPQA-D benchmark (+2.4\%). However, the introduction of a reward model incurs additional latency for scoring, which is a trade-off users must consider.

\paragraph{Combining with Length-Controlled Scaling.}
We also explored combining SeerSC with TokenSkip \citep{xia2025tokenskip}, a sequential scaling method designed to reduce token usage for reasoning models. We used a Qwen2.5-7B-Instruct model fine-tuned with TokenSkip (length parameter 0.7).

As shown in Table~\ref{tab:bon_tokenskip}(b), combining SeerSC with TokenSkip significantly reduces token consumption (up to 16\% on GSM8K). This comes with a slight accuracy drop, which is primarily attributable to the performance degradation inherent in the TokenSkip fine-tuning process.

\end{document}